\documentclass[conference]{IEEEtran}
\usepackage{graphicx}
\ifCLASSINFOpdf
\else
\fi
\usepackage[utf8]{inputenc}
\usepackage[T1]{fontenc}
\usepackage{amssymb}

\usepackage{cite}
\usepackage{amsmath}

\usepackage[numbers,sort&compress]{natbib}

\begin{document}
%
\title{Semi-Supervised Instance Population of an Ontology using Word Vector Embeddings}

\author{\IEEEauthorblockN{Vindula Jayawardana\IEEEauthorrefmark{1}, Dimuthu Lakmal\IEEEauthorrefmark{1}, Nisansa de Silva\IEEEauthorrefmark{1}, Amal Shehan Perera\IEEEauthorrefmark{1},\\Keet Sugathadasa\IEEEauthorrefmark{1}, Buddhi Ayesha\IEEEauthorrefmark{1}, Madhavi Perera\IEEEauthorrefmark{2}}
\IEEEauthorblockA{\IEEEauthorrefmark{1}Department of Computer Science \& Engineering\\University of Moratuwa}
\IEEEauthorblockA{\IEEEauthorrefmark{2}University of London International Programmes\\University of London}
\IEEEauthorblockA{Email: vindula.13@cse.mrt.ac.lk}
}


%


\maketitle

\begin{abstract}
In many modern day systems such as information extraction and knowledge management agents, ontologies play a vital role in maintaining the concept hierarchies of the selected domain. However, ontology population has become a problematic process due to its nature of heavy coupling with manual human intervention. With the use of word embeddings in the filed of natural language processing, it became a popular topic due to its ability to cope up with semantic sensitivity. Hence, in this study we propose a novel way of semi-supervised ontology population through word embeddings as the basis. We built several models including traditional benchmark models and new types of models which are based on word embeddings. Finally, we ensemble them together to come up with a synergistic model with better accuracy. We demonstrate that our ensemble model can outperform the individual models. \\
\end{abstract}


%
\IEEEpeerreviewmaketitle

\noindent\textbf{\textit{keywords}}: Ontology, Ontology Population, Word Embeddings, word2vec

\section{Introduction}
In various computational tasks in many different fields, the use of ontologies is becoming increasingly involved. Many of the research areas such as knowledge engineering and representation, information retrieval and extraction, and knowledge management and agent systems~\cite{Guarino} have incorporated the use of ontologies to a greater extent. As defined by Thomas R. Gruber~\cite{Gruber1993}, an ontology is a ``formal and explicit specification of a shared conceptualization''. Due to the evolving ability of ontologies to overcome limitations in traditional natural language processing methods, the popularity of using ontologies in modern computation tasks are getting increased day by day. For an example, text classification~\cite{textClassificationwithOntology,de2015safs3}, word set expansions~\cite{de2013semi}, linguistic information management~\cite{miller1990introduction,miller1990nouns,fellbaum1998wordnet, wijesiri2014building}, medical information management~\cite{huang2016omnisearch, huang2016development}, and information extraction~\cite{wimalasuriya2010ontology,de2017Discovering} emphasize the growing popularity of the ontology based computations and processing.

According to Carla Faria et al.~\cite{ontopop}, ontology population looks for instantiating the constituent elements of an ontology, like properties and non-taxonomic relationships. However, most of the time, ontology populations are done by domain experts and knowledge engineers as a manual process, which is both time consuming and expensive. As majority of the world’s knowledge is encoded in natural language text, automating the population of these ontologies using results obtained from Natural Language Processing (NLP) based analysis of documents, has recently become a major challenge for NLP applications~\cite{whyontoinNLP}.

In this study, we propose a novel way for semi-supervised instance population of an ontology using word vector embeddings. Word Embeddings could be identified as a collective name for a set of language modeling and feature learning techniques in natural language processing. The basic idea behind word embedding is based on the concept where words or phrases from the vocabulary are mapped to vectors of real numbers. We use these vectors as a method of arriving at instance population in an ontology. For this purpose, we built an iterative model based on the class representative vector for ontology classes~\cite{jayawardana2017deriving}. In our implementation, we built multiple models based on different methodologies. In one model we assigned membership to natural language tokens by distance to the representative vectors. In another, we used word2Vec’s internal dissimilar exclusion method to identify the membership. In another model, we used set expansion as described by~\cite{de2013semi}, for the purpose of ontology population. As each model outputs a set of candidate words for a given class, we then collaborate with domain experts and knowledge engineers to identify the performance of each model.

Semi-supervised learning falls between unsupervised learning (without any labeled training data) and supervised learning (with completely labeled training data). It has been observed that many machine leaning approaches elucidate considerable improvement in learning accuracy, when unlabeled data is used in conjunction with a small amount of labeled data.

The legal context contains jargon which is complex and most of the time impossible to have stored in mind; whether it be an average person or a paralegal, given that it consists terminology derived from ancient Latin terms, as well as various distinctive terminology depending on the category of laws and the geographical settings of practice. Therefore, knowing them manually is rather an impossible task which drove us to select the legal domain for this study of semi-supervised ontology population.

The rest of this paper is organized as follows: In Section~\ref{Background} we review previous studies related to this work. The details of our methodology for semi-supervised instance population of an ontology using word vector embeddings is introduced in Section~\ref{Methodology}. In Section~\ref{Results}, we demonstrate that our proposed methodology produces superior results outperforming traditional approaches. Finally, we conclude and discuss some future works in Section~\ref{conclusion}.

\section{Background and Related Work}
\label{Background}
The following sections depict the background of this study and other related studies. 


\subsection{Ontologies}
\label{Method:ontology}
Ontologies are mainly used to organize information as a form of knowledge representation in many areas. As defined by Thomas R. Gruber~\cite{Gruber1993}, 'ontologies are an explicit and formal specifications of the terms in the domain and the relations among them'. Ontologies have been expanding out from the realm of Artificial-Intelligence to domain specific tasks such as: Linguistics~\cite{wijesiri2014building,de2015safs3,de2013semi,de2011semap,de2017subject}, Law~\cite{jayawardana2017deriving}, Medicine~\cite{huang2016omnisearch,huang2016development,de2017Discovering}. Ontologies have become common on the semantic iteration of the World-Wide Web. An ontology may model either the world or a part of it as seen by the said area's viewpoint~\cite{de2013semi}.  

The basic ground units of an ontology are the \textit{Individuals} (instances). By grouping these \textit{Individuals} which can either be concrete objects or abstract objects, the structures called \textit{classes} are  built. A \textit{class} in an ontology is a representation of a concept, type, category, or a kind. However, these definitions may be altered depending on the domain of the ontology. Often these \textit{classes} form taxonomic hierarchies among them by subsuming, or being subsumed by, another class.

\subsection{Word Vector Embeddings}
\label{Method:wrd}

As first proposed by Tomas Mikolov et al.~\cite{mikolov2013distributed},  word embedding systems, are a set of natural language modeling and feature learning techniques, where words from a domain are mapped to vectors to create a model that has a distributed representation of words. Word2vec\footnote{https://code.google.com/p/word2vec/}\cite{mikolov2013efficient}, GloVe~\cite{pennington2014glove}, and Latent Dirichlet Allocation (LDA)~\cite{das2015gaussian} are the leading Word Vector Embedding systems. However, due to the flexibility and ease of customization, we picked word2vec as the word embedding method for this study.

Word2vec has been used in many areas due to its capability in coping up with the challenge of preserving the semantic sensitivity of a given context. It has been used in sentiment analysis~\cite{TwitterSentimental,sinaWeibo,chinessword2vec,sentimentalAnalysisInword2vec} and text classification~\cite{lilleberg2015support}. Gerhard Wohlgenannt et al.~\cite{word2VecOntologyPaper}'s approach to emulate a simple ontology using word2vec and Harmen Prins~\cite{word2VecOntologyPaper2}'s usage of word2vec extension: node2vec~\cite{node2VecOntologyPaper}, to overcome the problems in vectorization of an ontology, are two major works that have been carried out in relation to ontologies with the use of word2vec. More recently there have been successful studies on using word2vec on the legal domain~\cite{sugathadasa2017synergistic,jayawardana2017deriving}.

\subsection{Word Set Expansion}
\label{Method:wse}

Word lists that contain closely related sets of words is a critical requirement in machine understanding and processing of natural languages. Creating and maintaining such closely related word lists is a complex process that requires human input and is carried out manually in the absence of tools~\cite{de2013semi}. The said word-lists usually contain words that are deemed to be homogeneous in the level of abstraction involved in the application. Thus, two words $W_1$ and $W_2$ might belong to a single word-list in one application, but belong to different word-lists in another application. This fuzzy definition and usage is what makes creation and maintenance of these word-lists a complex task.    

De Silva et al.~\cite{de2013semi} describe a supervised learning mechanism which employs a word ontology to expand word lists containing closely related sets of words. This study has been an extension of their previous work~\cite{de2011semap}, which was done to enhance the refactoring process of the RelEx2Frame component of OpenCog AGI Framework, by expanding concept variables used in RelEx.

\subsection{Ontology Population}
\label{Method:ontoPop}

Being a knowledge acquisition task, ontology population is inherently a complex activity. Ontology population has been approached by using techniques such as rule based and machine learning. SPRAT~\cite{sprat} combines aspects from traditional named entity recognition, ontology-based information extraction, and relation extraction, in order to identify patterns for the extraction of a variety of entity types and relations between them, and to
re-engineer them into concepts and instances in an ontology. Rene Witte et al.~\cite{whyontoinNLP} has developed a GATE resource called the OwlExporter, that allows to easily map existing NLP analysis pipelines to OWL ontologies, thereby allowing language engineers to create ontology population systems without requiring extensive knowledge of ontology APIs. 

However modern day recherches are more focused on semi supervised ontology population due to the nature of less manual intervention. 

\subsection{Semi Supervised Ontology Population}
\label{Method:semiOntoPop}

Although supervised machine learning methodologies have showed promising results when it comes to information extraction, they accumulate more cost for training since they require vast number of labeled training data. As a solution, semi-supervised machine learning methodologies have been introduced, requiring considerably less amount of labeled training data.

Carlson~\cite{calson2010paper} proposed a semi-supervised learning model to populate instances of a set of target categories and relations of an ontology by providing seed labeled data and a set of constraints which couples classes and relationships of an ontology. Semi-supervised algorithms tend to show unacceptable results due to ‘semantic drift’ and constraints have been introduced to overcome the issue. Carlson has used ‘Bootstrapping’ method for semi-supervised learning which starts with a small number of labeled data and grows labeled data iteratively, which are chosen from a set of candidates, which is classified using the current semi-supervised model. Three types of constraints have been introduced by Carlson to conform mutual exclusion, type checking, and text features.

Carlson~\cite{calsonnellpaper} has expanded coupled semi-supervised learning~\cite{calson2010paper} to never-ending language learning (NELL); an agent that runs forever to extract information from the web and populate them continuously into a knowledge base. A prototype of the system that they have implemented is able to extract noun phrases related to various semantic categories, and semantic relations between categories. Its information extracting ability increases day by day which is evidenced by the ability to extract more information from previous day's text sources more accurately. Input ontology in the system was included with seed instances for each ontology class and then sub systems which consist of previously described coupled semi supervised methodologies extract candidate instances and relationships from the text corpus. Knowledge Integrator of the system choose strongly supported sets of instances and relations from the candidate set, as new beliefs of the system.

Zhilin Yang~\cite{graphembeddingspaper}  has presented a semi supervised learning methodology based on graph embeddings. The system consists of two main sections namely ‘transductive’ and ‘inductive’. The ‘transductive’ approach predicts instances which are already observed in the graph in the training period. In ‘inductive’ approach, predictions can be made on unobserved instances in the training period. A probabilistic model was developed to learn node embeddings to generate edges in a graph.




\section{Methodology}
\label{Methodology}

\begin{figure*}[!htb]	
\centering
\includegraphics[width=0.75\textwidth]{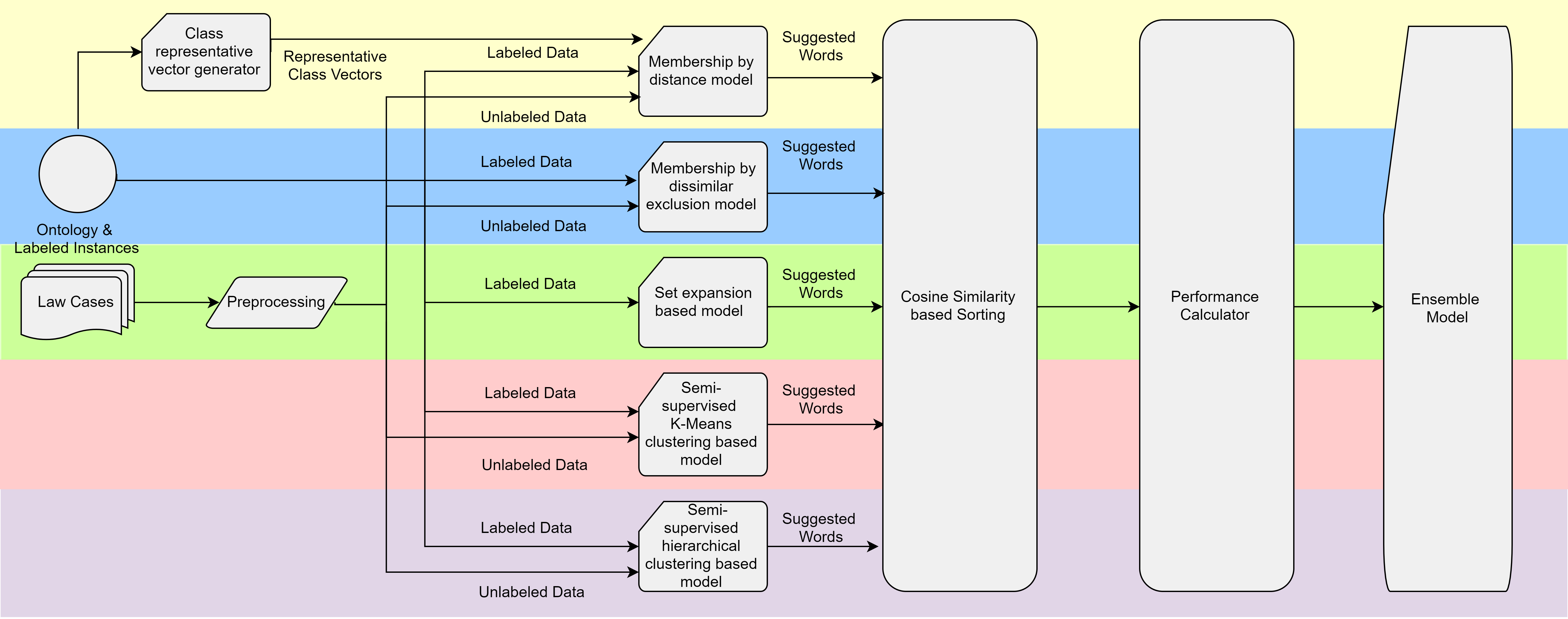}
\caption{Flow of Semi-Supervised Instance Population of an Ontology using Word Vector Embeddings }	
\label{fig:flowDiagram}
\end{figure*}

We discuss the methodology used in this research study in the upcoming sections. Each of the following subsections describe a step of our process. An overview of the methodology we propose is illustrated in~\figurename~\ref{fig:flowDiagram}.

\subsection{Ontology Creation}
\label{Method:onto}

\begin{figure}[!hb]
\includegraphics[width=0.5\textwidth]{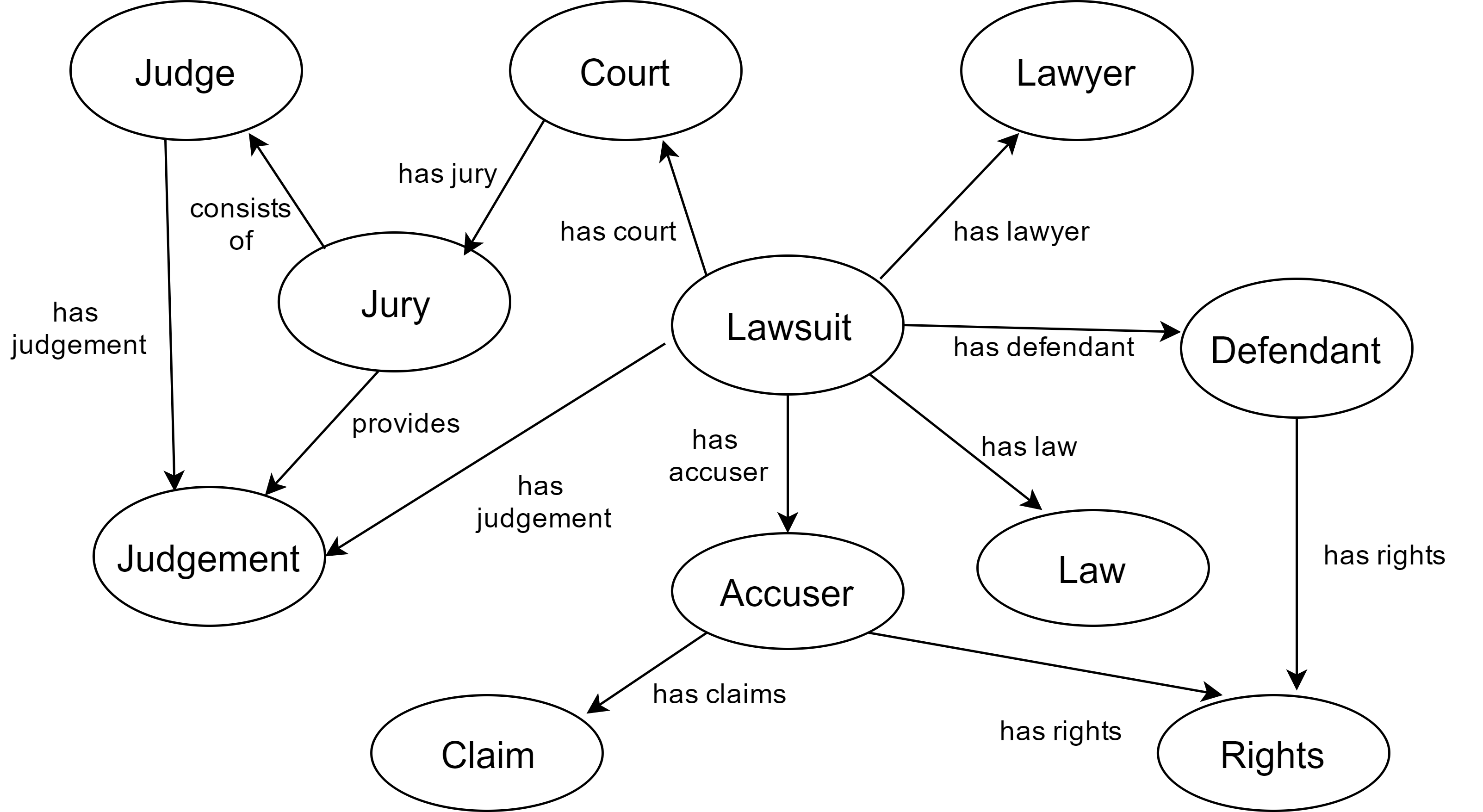}
\caption{Ontology Sub-section used for the Population}
\label{fig:ontology}
\end{figure}

For the ontology creation, we focused on the consumer protection law domain and created a legal ontology, based on Findlaw~\cite{caseLaw} as the reference. However, for the sake of clarity of this paper, we extract a sub-ontology from it and use it to explain the methodology to make the process simple and intuitive to understand. In selecting a part of the ontology, we mainly focused on more sophisticated relationships and taxonomic presences. An overview of the selected part of ontology is illustrated in~\figurename~\ref{fig:ontology}. After the creation of sub-ontology, we manually populated the ontology with seed instances with collaboration of domain experts and knowledge engineers.

\subsection{Training word Embeddings}
\label{Method:wordEm}
The word embeddings method used in this study was built using a \textit{word2vec} model. We obtained a large legal text corpus from Findlaw~\cite{caseLaw} and built a \textit{word2vec} model using the corpus. The reason for selecting word2vec word embedding for this study is the success demonstrated by other studies such as~\cite{jayawardana2017deriving} and~\cite{sugathadasa2017synergistic} in the legal domain that uses word2vec as the word embedding method. The text corpus consisted of legal cases under 78 law categories. In creating the legal text corpus, we used the \textit{Stanford CoreNLP} for preprocessing the text with tokenizing and sentence splitting. Following are the important parameters we specified in training the model.

\begin{itemize}
  \item size (dimensionality): 200 
  \item context window size: 10
  \item learning model: CBOW
  \item min-count: 5
  \item training algorithm: hierarchical softmax
\end{itemize}

\subsection{Deriving Representative Class Vectors}
\label{Method:ClassVectors}
Ontology classes are sets of homogeneous instance objects that can be converted to a vector space by word vector embeddings. A methodology to derive a representative vector for ontology classes, whose instances were mapped to a vector space is presented in~\cite{jayawardana2017deriving}. We followed the same approach and started by deriving five candidate vectors which are then used to train a machine learning model that would calculate a representative vector for each of the classes in the selected sub-ontology shown in~\figurename~\ref{fig:ontology}. In the following sections, we describe in-depth, the manner in how we used this derived class vectors in our proposed methodology.

\subsection{Instance Corpus for Ontology Population}
\label{Method:Corpus}
In order to perform semi-supervised ontology population, we used legal cases from Findlaw~\cite{caseLaw} to create an instance corpus. We performed \textit{Stanford CoreNLP} based preprocessing on the raw text with tokenizing and sentence splitting to generate the instance corpus. This legal corpus was used in the subsequent models for the purpose of ontology population.
 
\subsection{Candidate Model Building}
\label{Method:Candidate}

Based on the aforementioned components, we built five candidate models for semi-supervised population of the ontology. The five models are illustrated below:

\begin{itemize}
\item Membership by distance model ($M_{1}$)
\item Membership by dissimilar exclusion model ($M_{2}$) 
\item Set expansion based model ($M_{3}$)
\item Semi-supervised K-Means clustering based model ($M_{4}$)
\item Semi-supervised hierarchical clustering based model  ($M_{5}$)
\end{itemize}
\hfill
\subsubsection{Membership by Distance Model  ($M_{1}$)}
In this model, the candidate vectors for the ontology were generated from the instance corpus based on the minimum distance to the representative class vector derived in Section~\ref{Method:ClassVectors}. In taking the vector similarity, we used cosine similarity. Given an instance $i$ which has the vector embedding $X_i$, Equation~\ref{Eq:VectorMedianFormula} describes which class the particular instance belongs to.
\begin{equation}
C_{M1} =\Bigg\{ j~\Bigg|~\underset{c_j \in C}{\mathrm{argmax}} \Bigg\{ \frac{X_{i}.c_j}{|X_{i}||c_j|}  \Bigg\} \Bigg\}
\label{Eq:VectorMedianFormula}
\end{equation}

Here, the set $C$ denotes the set of representative class vectors. $C_{M1}$ is the selected class index of the instance $i$ out of class set $C$. 

\subsubsection{Membership by dissimilar exclusion model ($M_{2}$)}
In this model, we used the word2vec based dissimilar exclusion method in identifying the membership of a particular instance to a given class. This is a utilization of an internal method of word2vec where given a set of members, it would return the member that should be removed from the set in order to increase the set cohesion. For example, given the set of instances: \textit{breakfast}, \textit{cereal}, \textit{dinner} and \textit{lunch}, the word2vec dissimilar exclusion method would identify the instance \textit{cereal} as the item that should be removed from the set to increase the set cohesion. We define this method as shown in Equation~\ref{Eq:Exclusion} where $S$ is the set provided and $e$ is the member selected to be excluded.   

\begin{equation}
e=Exclusion(S)
\label{Eq:Exclusion}
\end{equation}

We used the Equation~\ref{Eq:ExclusionSet} to decide whether the instance $i$ should belong to class $j$. Here, $S_j$ is the seed set of class $j$ and $X_i$ is the vector representation of the instance $i$. If the value $E_{i,j}$ gets evaluated to $TRUE$ we declare that instance $i$ should belong to class $j$ under model $M_2$.     

\begin{equation}
E_{i,j}=\bigg\{ e \in S_j \bigg| e=Exclusion(S_j \cap X_i) \bigg\}
\label{Eq:ExclusionSet}
\end{equation}

When using the aforementioned method in identifying the membership of an instance, there is a possibility of getting more than one class for a given instance as a possible parent class. Hence; 

\begin{equation}
C_{M2} = \Big\{C_{k}~\Big|~0 < k \leqslant N \Big\}
\label{Eq:candidateMatrix}
\end{equation}

Here in Equation~\ref{Eq:candidateMatrix}, $C_{M2}$ is the set of classes for a given instance $i$. $C_k$ denote the common representation of those set of classes and $N$ denotes the total number of classes we have in the ontology.

\subsubsection{Set Expansion Based Model ($M_{3}$)}
For the purpose of set expansion based model, we selected the algorithm presented in~\cite{de2013semi}, which was built on the earlier algorithm described in~\cite{de2011semap}. The rationale behind this selection is the fact that as per~\cite{de2013semi},  WordNet~\cite{miller1990introduction} based linguistic processes are reliable due to the fact that the WordNet lexicon was built on the knowledge of expert linguists.

In this model, the idea is to increase the ontology class instances based on a WordNet hierarchy based expansion. Simply put, it discovers the WordNet $synsets$ pertaining to the seed words and proceeds up the hierarchy to find the minimum common ancestors for each of the senses of the words. Next, the most common word sense is selected by majority. The relevant rooted tree is extracted and the gazetteer list of that rooted synset tree is created. The gazetteer list is subjected to a set subtraction of the original seed set. The set intersection of the remaining set with the candidate word set is declared to be the word set assigned to the given class. However, it should be noted that as we showed in model $M_2$, after running the set expansion algorithm, one candidate instance may be tentatively assigned to more than one class.


\subsubsection{Semi-Supervised K-Means Clustering Based Model ($M_{4}$)}
\label{Method:KM}
Out of the models proposed in this study so far, this model is the first semi-supervised model. First, the seed instances are put together with the unlabeled data from instance corpus. Let $N_{labeled}$ be the number of labeled (seed) instances and $N_{unlabeled}$ be the total number of unlabeled instances. Thus, by mixing up the labeled and unlabeled data, we get a total of $N_{labeled}$ + $N_{unlabeled}$ number of instances. Next all the instances are subjected to the k-means algorithm where $k$ is selected to the the same, as the number of classes in the ontology.




\begin{figure}[!htb]
\centering
\includegraphics[width=0.2\textwidth]{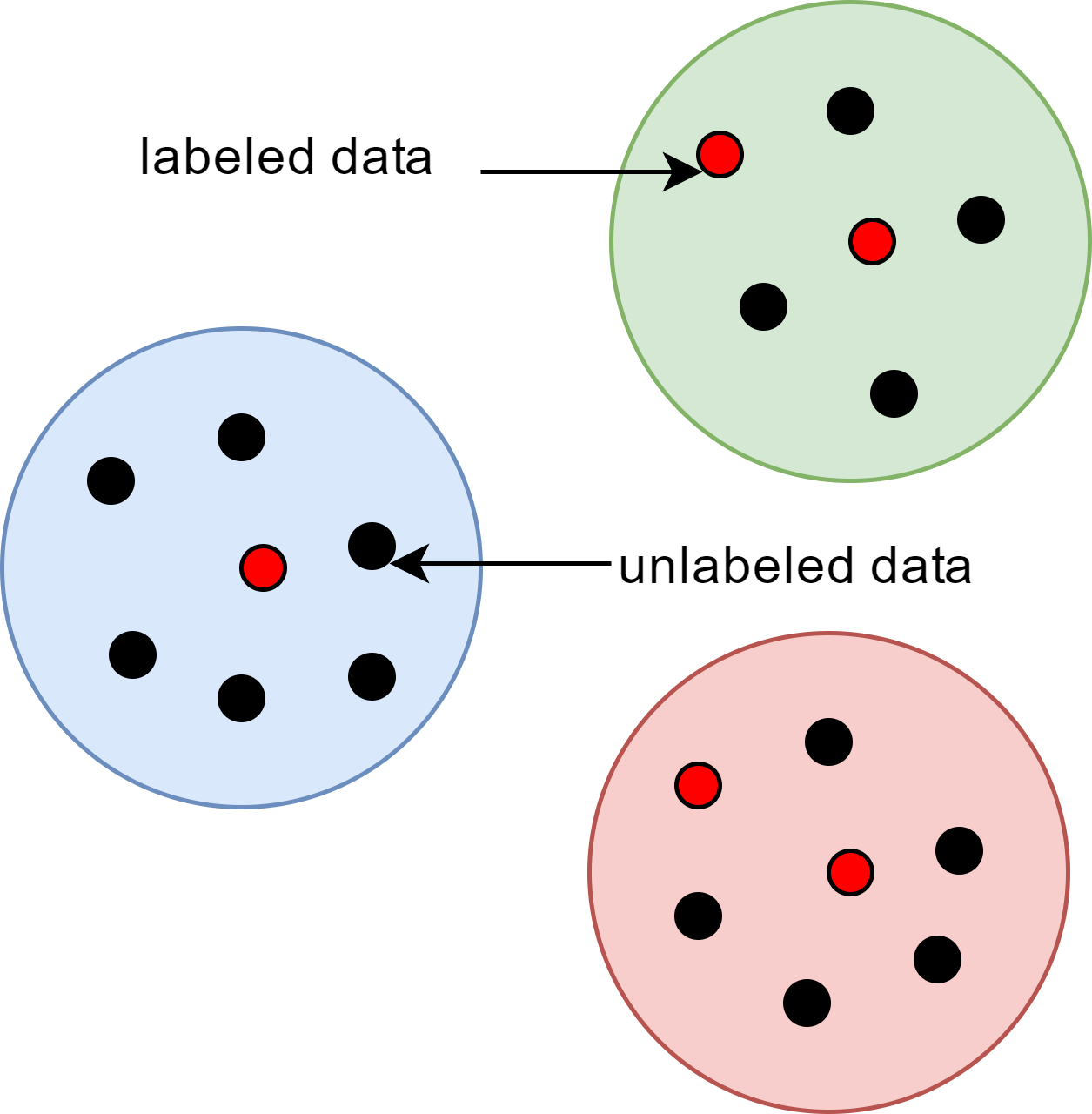}
\caption{An illustration of k-means clustering with k = 3.}
\label{fig:cluster}
\end{figure}

Once the k-means clustering is finished, primary class cluster assignment for cluster $L$ is done by voting of seed instances according to Equation~\ref{Eq:voteMax}, where $C$ is the set of ontology classes, $c_j$ is the $j$th class from $C$, $y_i$ is the $i$th instance from $L$, and $d_i$ is defined according to Equation~\ref{Eq:DefineDi}.  


\begin{equation}
C_{l} =\Bigg\{ j~\Bigg|~\underset{c_j \in C}{\mathrm{argmax}} \Big\{  ~\underset{y_i \in L}{\sum d_i} \Big\}   \Bigg\}
\label{Eq:voteMax}
\end{equation}


\begin{equation}
d_i=
\begin{cases}
    1 & \text{if } y_i \in c_j \\
    0               & \text{otherwise}
\end{cases}
\label{Eq:DefineDi}
\end{equation}

At this point, it should be noted that there can be three situations where it is possible to not get a $c_l$ value assigned to some class $L$ by Equation~\ref{Eq:voteMax} without ambiguity: (1) $L$ not having any seed instances to vote. (2) $L$ has multiple seed instances but the majority voting ended in a tie. (3) Two (or more) clusters, claim the same class. To solve these problems we defined Equation~\ref{Eq:VectorUnassigned}, which selected the unassigned class that is closest to an unassigned cluster. Here, an unassigned cluster $L'$ is considered. $C'$ is the set of representative class vectors of unassigned classes. $C_{l'}$ is the selected class index of the cluster $L'$.

\begin{equation}
C_{l'} =\Bigg\{ j~\Bigg|~\underset{c_j \in C'}{\mathrm{argmax}} \Bigg\{ \underset{x_i \in L'}{\sum} \Bigg\{  \frac{X_{i}.c_j}{|X_{i}||c_j|} \Bigg\} \Bigg\} \Bigg\}
\label{Eq:VectorUnassigned}
\end{equation}

The first problem to be solved is the problem of $L$ having multiple seed instances but the majority voting ending in a tie. In this case the $C'$ of Equation~\ref{Eq:VectorUnassigned} is limited to the set intersection of tied classes and unassigned classes. Next, the problem of Two (or more) clusters, claiming the same class is solved. In this case $C'$ of Equation~\ref{Eq:VectorUnassigned} is limited to the contested class. These steps are repeated until there is an iteration where there are no new assignments. Finally, all the remaining unassigned classes are put in $c'$ and Equation~\ref{Eq:VectorUnassigned} is executed repetitively with tie breaking, done with precedence until all the clusters are uniquely assigned to some class.





\subsubsection{Semi-Supervised Hierarchical Clustering Based Model ($M_{5}$)}
The next model being used is a semi-supervised method based on hierarchical clustering. Hierarchical clustering is a method of cluster analysis which seeks to build a hierarchy of clusters. We built a model which creates such hierarchy of clusters using the word embeddings taken from the word2vec model,of the entire corpus similar to the process in Section~\ref{Method:KM}. In this model, we extracted the slice of hierarchical clusters such that the number of clusters in the slice is equal to the number of classes in the sub-ontology. Next, the cluster-class assignment was done similar to the process in~\ref{Method:KM}. 


\subsection{Model Accuracy Measure}
\label{accuracyMeasure}
After building the aforementioned models, we evaluated the accuracy of each model. As each model outputs an unordered set of suggested words, we sorted them using the Neural Network, trained according to the methodology proposed in~\cite{sugathadasa2017synergistic}. Upon completing the sorting, we applied a threshold to select the best candidates. Finally, we measured each model's accuracy as below. For this task, we involved with domain experts and knowledge engineers. For a given model $M_i$ in the context of class $j$: 

\begin{equation}
Precision_{M_{i,j}}= 
\frac{W_{M_{i,j}} \cap W_{j,g}}{W_{M_{i,j}}}
\label{Eq:precision}
\end{equation}

\begin{equation}
Recall_{M_{i,j}}= 
\frac{W_{M_{i,j}} \cap W_{j,g}}{W_{j,g}}
\label{Eq:recall}
\end{equation}

Here, $W_{M_i}$ denotes words by the model $M_i$ and $W_{j,g}$ denotes the set of the words proposed by domain experts that needs to be the golden standard for class $j$. The model precision and model recall of $M_i$ was calculated by averaging the class values for precision and recall of those models. 


\begin{equation}
F1_{M_{i}}= 2.
\frac{Precision_{M_{i}}.Recall_{M_{i}}}{Precision_{M_{i}} + Recall_{M_{i}}}
\label{Eq:F1}
\end{equation}

\subsection{Ensemble Model}
\label{ensembelModel}
Next, we came up with an ensemble model based on the models identified earlier. In the task of creating the  ensemble model, we allocated a candidate weight for each model based on each model's $F1$ measure as calculated in the previous step. 

Let $M_i$ be a model, out of the obtained models, and let $F1_{M_{i}}$ be the F1 measure of model $M_i$. Hence with the models in consideration, weight of the model $W_i$ is calculated as shown in equation~\ref{Eq:VectorAverageFormula}, where $p$ is the total number of models.

\begin{equation}
W_i = \frac{F1_i }{\displaystyle \sum_{i=1}^{p} F1_i }
\label{Eq:VectorAverageFormula}
\end{equation}

As identified above, upon calculating the weight of each model, we created the ensemble model as shown in equation~\ref{Eq:modelMatrix}. Given an unlabeled instance $Y$, let $M_{ensemble}$ be a $p \times n$ matrix where $n$ denotes the number of classes in the ontology and $p$ denotes the number of basic models. Each column of the matrix corresponds to a class in the ontology and each row corresponds to a model, while each $m_{i,j}$ is derived from Equation~\ref{Eq:mi}.

\begin{equation}
M_{ensemble} = 
\begin{bmatrix}
   m_{1,1} & m_{1,2} & \dots & m_{1,n} \\
   m_{2,1} & m_{2,2} & \dots & m_{2,n} \\
   m_{3,1} & m_{3,2} & \dots & m_{3,n} \\
   \dots & \dots & \dots & \dots \\
   m_{p,1} & m_{p,2} & \dots & m_{p,n} \\
\end{bmatrix}
\label{Eq:modelMatrix}
\end{equation}

\begin{equation}
m_{i,j} = 
\begin{cases}
    1 & \text{if } Y \in M_i \\
    0               & \text{otherwise}
\end{cases}
\label{Eq:mi}
\end{equation}

Let$M_{weights}$ be the $p$ length vector which defines the weights of each model calculated by equation~\ref{Eq:VectorAverageFormula}.

\begin{equation}
M_{weights} = 
\begin{bmatrix}
   w_1 & w_2 & w_3 & \dots & w_p\\
\end{bmatrix}
\label{Eq:weightMatrix}
\end{equation}

Then we calculate the total score vector for the instance $Y$ by, 

\begin{equation}
S = M_{weights}.M_{ensemble}
\label{Eq:score}
\end{equation}

Here, $S$ is the score vector of size $n$, where element $i$ in the vector denotes the total score for instance $Y$ for the membership in Class $C_{i}$. Next, we selected the class with the highest membership score as the parent class of instance $Y$. It is illustrated in Equation~\ref{Eq:voteMaxmember}.

\begin{equation}
C_{M_{ensemble}} =\bigg\{~i~\bigg|~\underset{S_{C_i} \in S}{\mathrm{argmax}} \Big\{ S_{C_i}\Big\} \bigg\}
\label{Eq:voteMaxmember}
\end{equation}

With that, we got the final class of the instance $Y$. Hence, we populate that selected class with the instance $Y$.

\section{Results}
\label{Results}

\begin{figure}[!htb]
\centering
\includegraphics[width=0.5\textwidth]{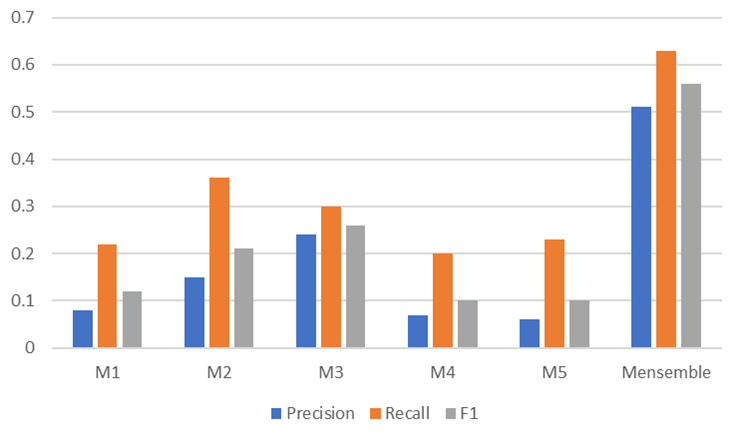}
\caption{Comparison of precision, recall and F1 of the models}
\label{fig:graph}
\end{figure}

In testing our ensemble model, we used another instance corpus. In this corpus, we subdivided in the order of 70\%, 20\%, and 10\% as the training set, the validation set, and the test set respectively. Training set was used in training the models individually. Validation set was used to fine tune the models. Finally, the test set was used in verifying the accuracy of the models. We report our findings below in the table~\ref{tab:res}, where we compare the individual models: membership by distance model($M_1$), membership by dissimilar exclusion model($M_2$), set expansion based model($M_3$), k-means clustering based model($M_4$), hierarchical clustering based model($M_5$) and the ensemble model as a whole. In Fig.~\ref{fig:graph}, we compare the precision, recall and F1 of each of the candidate models along with the ensemble model.  

\begin{table}[!htb]
	\centering
	\caption{Comparison of performance of models}\label{tab:res}
	\begin{tabular}{|l|c|c|c|}
    	\hline
        & Precision & Recall & F1\\
		\hline
		$M_1$ & 0.08 & 0.22 & 0.12 \\
        \hline
        $M_2$ & 0.15 & 0.36 & 0.21\\
        \hline
        $M_3$ & 0.24 & 0.30 & 0.26\\
        \hline
        $M_4$ & 0.07 & 0.20 & 0.10\\
        \hline
        $M_5$ & 0.06 & 0.23 & 0.10\\
        \hline
        \hline
        $M_{ensemble}$ & \textbf{0.51} & \textbf{0.63} & \textbf{0.56}\\
        \hline
	\end{tabular}	
\end{table}

In defining the ensemble model, Equation~\ref{Eq:wei} defines the calculated weights of each model in the order of models M1 to M5. These calculations are based on the above mentioned training set.
\begin{equation}
M_{weights} = 
\begin{bmatrix}
   0.15 & 0.27 & 0.33 & 0.13 & 0.12\\
\end{bmatrix}
\label{Eq:wei}
\end{equation}

As can be seen from Table~\ref{tab:res}, our ensemble model's F1 has been improved by 0.30, compared to the best of the candidate models. Hence, from the results obtained, as a proof of concept, we can demonstrate that word embeddings can be used effectively in semi-supervised ontology population.

\section{Conclusion and Future Works}
\label{conclusion}
Through this work we demonstrated the use of word embeddings on semi-supervised ontology population. We mainly focused on semi-supervised population which basically falls between the supervised population and unsupervised population. The main motive behind making the process semi-supervised is to reduce the level of manual interventions in ontology populations while maintaining a considerable amount of accuracy. As shown in the results, our ensemble model outperforms the five individual models in populating the selected legal ontology. The findings in this study is mainly important in two ways as mentioned below.

Firstly, an important part of the ontology engineering cycle is the ability to keep a handcrafted ontology up to date. Through the semi-supervised ontology population we can reduce the hassle involved in manual intervention to keep the ontology updated. 

Secondly, there is novelty in the methodology proposed in our study. We proved that, since word embeddings map words or phrases from the vocabulary to vectors of real numbers based on the semantic context, a methodology based upon it can yield more sophisticated results when it comes to context sensitive tasks like ontology population. This indeed is a step up from the traditional information extraction based ontology population and maintenance processes, towards new horizons. 

We can improve the methodology proposed, to yield better accuracy performances. For an example, we only considered the single word instances in populating the ontology using the defined models. However, in some of the scenarios, phrases also could be instances of ontology classes. Hence, it is important to convert phases to vectors and use them in the methodology as well. Also, as illustrated with models $M_4$ and $M_5$, we can perform more sophisticated semi-supervised ontology populations based on the concept of this study with more improvements. We keep them to be the future works of this study.

\bibliographystyle{IEEEtran}
\bibliography{VectorRep}

\begin{thebibliography}{10}
\providecommand{\url}[1]{#1}
\csname url@samestyle\endcsname
\providecommand{\newblock}{\relax}
\providecommand{\bibinfo}[2]{#2}
\providecommand{\BIBentrySTDinterwordspacing}{\spaceskip=0pt\relax}
\providecommand{\BIBentryALTinterwordstretchfactor}{4}
\providecommand{\BIBentryALTinterwordspacing}{\spaceskip=\fontdimen2\font plus
\BIBentryALTinterwordstretchfactor\fontdimen3\font minus
  \fontdimen4\font\relax}
\providecommand{\BIBforeignlanguage}[2]{{%
\expandafter\ifx\csname l@#1\endcsname\relax
\typeout{** WARNING: IEEEtran.bst: No hyphenation pattern has been}%
\typeout{** loaded for the language `#1'. Using the pattern for}%
\typeout{** the default language instead.}%
\else
\language=\csname l@#1\endcsname
\fi
#2}}
\providecommand{\BIBdecl}{\relax}
\BIBdecl

\bibitem{Guarino}
N.~Guarino, ``Formal ontology and information systems,'' \emph{Proceedings of
  FOIS’98, Trento, Italy}, 1998.

\bibitem{Gruber1993}
T.~R. Gruber, ``A translation approach to portable ontology specifications,''
  \emph{Knowledge Acquisition, 5(2):199-220}, 1993.

\bibitem{textClassificationwithOntology}
X.-Q. Yang, N.~Sun, T.-L. Sun \emph{et~al.}, ``The application of latent
  semantic indexing and ontology in text classification,'' \emph{International
  Journal of Innovative Computing, Information and Control}, vol.~5, no.~12,
  pp. 4491--4499, 2009.

\bibitem{de2015safs3}
N.~de~Silva, ``Safs3 algorithm: Frequency statistic and semantic similarity
  based semantic classification use case,'' \emph{Advances in ICT for Emerging
  Regions (ICTer), 2015 Fifteenth International Conference on}, pp. 77--83,
  2015.

\bibitem{de2013semi}
N.~De~Silva, A.~Perera, and M.~Maldeniya, ``Semi-supervised algorithm for
  concept ontology based word set expansion,'' \emph{Advances in ICT for
  Emerging Regions (ICTer), 2013 International Conference on}, pp. 125--131,
  2013.

\bibitem{miller1990introduction}
G.~A. Miller, R.~Beckwith, C.~Fellbaum, D.~Gross, and K.~J. Miller,
  ``Introduction to wordnet: An on-line lexical database,'' \emph{International
  journal of lexicography}, vol.~3, no.~4, pp. 235--244, 1990.

\bibitem{miller1990nouns}
G.~A. Miller, ``Nouns in wordnet: a lexical inheritance system,''
  \emph{International journal of Lexicography}, vol.~3, no.~4, pp. 245--264,
  1990.

\bibitem{fellbaum1998wordnet}
C.~Fellbaum, \emph{WordNet}.\hskip 1em plus 0.5em minus 0.4em\relax Wiley
  Online Library, 1998.

\bibitem{wijesiri2014building}
I.~Wijesiri, M.~Gallage, B.~Gunathilaka, M.~Lakjeewa, D.~C. Wimalasuriya,
  G.~Dias, R.~Paranavithana, and N.~De~Silva, ``Building a wordnet for
  sinhala,'' in \emph{7th Global Wordnet Conference}, 2014, p. 100.

\bibitem{huang2016omnisearch}
J.~Huang, F.~Gutierrez, H.~J. Strachan, D.~Dou, W.~Huang, B.~Smith, J.~A.
  Blake, K.~Eilbeck, D.~A. Natale, Y.~Lin \emph{et~al.}, ``Omnisearch: a
  semantic search system based on the ontology for microrna target (omit) for
  microrna-target gene interaction data,'' \emph{Journal of biomedical
  semantics}, vol.~7, no.~1, p.~1, 2016.

\bibitem{huang2016development}
J.~Huang, K.~Eilbeck, B.~Smith, J.~A. Blake, D.~Dou, W.~Huang, D.~A. Natale,
  A.~Ruttenberg, J.~Huan, M.~T. Zimmermann \emph{et~al.}, ``The development of
  non-coding rna ontology,'' \emph{International journal of data mining and
  bioinformatics}, vol.~15, no.~3, pp. 214--232, 2016.

\bibitem{wimalasuriya2010ontology}
D.~C. Wimalasuriya and D.~Dou, ``Ontology-based information extraction: An
  introduction and a survey of current approaches,'' \emph{Journal of
  Information Science}, 2010.

\bibitem{de2017Discovering}
N.~de~Silva, D.~Dou, and J.~Huang, ``Discovering inconsistencies in pubmed
  abstracts through ontology-based information extraction,'' in
  \emph{Proceedings of the 8th ACM International Conference on Bioinformatics,
  Computational Biology, and Health Informatics}.\hskip 1em plus 0.5em minus
  0.4em\relax ACM, 2017, pp. 362--371.

\bibitem{ontopop}
R.~G. Carla~Fariaa, Ivo~Serrab, ``A domain-independent process for automatic
  ontology population from text,'' \emph{Science of Computer Programming},
  2014.

\bibitem{whyontoinNLP}
J.~R. Rene~Witte, Ninus~Khamis, ``Flexible ontology population from text: The
  owlexporter.''

\bibitem{jayawardana2017deriving}
V.~Jayawardana, D.~Lakmal, N.~de~Silva, A.~S. Perera, K.~Sugathadasa, and
  B.~Ayesha, ``Deriving a representative vector for ontology classes with
  instance word vector embeddings,'' \emph{arXiv preprint arXiv:1706.02909},
  2017.

\bibitem{de2011semap}
N.~de~Silva, C.~Fernando, M.~Maldeniya, D.~Wijeratne, A.~Perera, and
  B.~Goertzel, ``Semap-mapping dependency relationships into semantic frame
  relationships,'' in \emph{17th ERU Research Symposium}, vol.~17.\hskip 1em
  plus 0.5em minus 0.4em\relax Faculty of Engineering, University of Moratuwa,
  Sri Lanka, 2011.

\bibitem{de2017subject}
N.~de~Silva, D.~Maldeniya, and C.~Wijeratne, ``Subject specific stream
  classification preprocessing algorithm for twitter data stream,'' \emph{arXiv
  preprint arXiv:1705.09995}, 2017.

\bibitem{mikolov2013distributed}
T.~Mikolov, I.~Sutskever, K.~Chen, G.~S. Corrado, and J.~Dean, ``Distributed
  representations of words and phrases and their compositionality,''
  \emph{Advances in neural information processing systems}, pp. 3111--3119,
  2013.

\bibitem{mikolov2013efficient}
T.~Mikolov, I.~Sutskever, K.~Chen, G.~Corrado, and J.~Dean, ``Efficient
  estimation of word representations in vector space,'' \emph{arXiv preprint
  arXiv:1301.3781}, 2013.

\bibitem{pennington2014glove}
J.~Pennington, R.~Socher, and C.~D. Manning, ``Glove: Global vectors for word
  representation.'' in \emph{EMNLP}, vol.~14, 2014, pp. 1532--1543.

\bibitem{das2015gaussian}
R.~Das, M.~Zaheer, and C.~Dyer, ``Gaussian lda for topic models with word
  embeddings.'' in \emph{ACL (1)}, 2015, pp. 795--804.

\bibitem{TwitterSentimental}
D.~Tang, F.~Wei, N.~Yang, M.~Zhou, T.~Liu, and B.~Qin, ``Learning
  sentiment-specific word embedding for twitter sentiment classification,''
  \emph{Proceedings of the 52nd Annual Meeting of the Association for
  Computational Linguistics}, vol.~1, pp. 1555--1565, 2014.

\bibitem{sinaWeibo}
B.~Xue, C.~Fu, and Z.~Shaobin, ``Study on sentiment computing and
  classification of sina weibo with word2vec,'' \emph{Big Data (BigData
  Congress), 2014 IEEE International Congress on. IEEE}, pp. 358--363, 2014.

\bibitem{chinessword2vec}
D.~Zhang, H.~Xu, Z.~Su, and Y.~Xu, ``Chinese comments sentiment classification
  based on word2vec and svm perf,'' \emph{Expert Systems with Applications},
  vol.~42, no.~4, pp. 1857--1863, 2015.

\bibitem{sentimentalAnalysisInword2vec}
H.~Liu, ``Sentiment analysis of citations using word2vec,'' \emph{arXiv
  preprint arXiv:1704.00177}, 2017.

\bibitem{lilleberg2015support}
J.~Lilleberg, Y.~Zhu, and Y.~Zhang, ``Support vector machines and word2vec for
  text classification with semantic features,'' in \emph{Cognitive Informatics
  \& Cognitive Computing (ICCI* CC), 2015 IEEE 14th International Conference
  on}.\hskip 1em plus 0.5em minus 0.4em\relax IEEE, 2015, pp. 136--140.

\bibitem{word2VecOntologyPaper}
G.~Wohlgenannt and F.~Minic. Using word2vec to build a simple ontology learning
  system. Available at: \url{http://ceur-ws.org/Vol-1690/paper37.pdf}.
  Accessed: 2017-05-30.

\bibitem{word2VecOntologyPaper2}
H.~Prins, ``Matching ontologies with distributed word embeddings.''

\bibitem{node2VecOntologyPaper}
A.~Grover and J.~Leskovec, ``node2vec: Scalable feature learning for
  networks,'' pp. 855--864, 2016.

\bibitem{sugathadasa2017synergistic}
K.~Sugathadasa, B.~Ayesha, N.~de~Silva, A.~S. Perera, V.~Jayawardana,
  D.~Lakmal, and M.~Perera, ``Synergistic union of word2vec and lexicon for
  domain specific semantic similarity,'' \emph{arXiv preprint
  arXiv:1706.01967}, 2017.

\bibitem{sprat}
A.~F. Diana~Maynard and W.~Peters, ``Sprat: a tool for automatic semantic
  pattern-based ontology population,'' 2009.

\bibitem{calson2010paper}
A.~Carlson, J.~Betteridge, R.~C. Wang, E.~R. Hruschka, Jr., and T.~M. Mitchell,
  ``Coupled semi-supervised learning for information extraction,'' \emph{WSDM
  '10 Proceedings of the third ACM international conference on Web search and
  data mining}, pp. 101--110, 2010.

\bibitem{calsonnellpaper}
A.~Carlson, J.~Betteridge, B.~Kisiel, B.~Settles, E.~R. Hruschka, Jr., and
  T.~M. Mitchell, ``Toward an architecture for never-ending language
  learning,'' \emph{Proceeding AAAI'10 Proceedings of the Twenty-Fourth AAAI
  Conference on Artificial Intelligence}, pp. 1306--1313, 2010.

\bibitem{graphembeddingspaper}
R.~S. Zhilin~Yang, William W.~Cohen, ``Revisiting semi-supervised learning with
  graph embeddings,'' \emph{Proceedings of International Conference on Machine
  Learning}, 2016.

\bibitem{caseLaw}
``{FindLaw} cases and codes,'' \url{http://caselaw.findlaw.com/}, accessed:
  2017-05-18.

\end{thebibliography}

\end{document}